\documentclass[conference]{IEEEtran}
\usepackage{cite}
\usepackage[pdftex]{graphicx}
\usepackage{array}
\usepackage{url}
\usepackage{cite}
\usepackage{tabularx,booktabs}
\usepackage{amsmath,amssymb,amsfonts}
\usepackage{textcomp}
\usepackage[flushleft]{threeparttable}
\usepackage{footnote}
\usepackage{setspace}
\usepackage{color}
\usepackage{algorithm}
\usepackage{algpseudocode}
\usepackage{multicol}

\ifCLASSOPTIONcompsoc
    \usepackage[caption=false, font=normalsize, labelfont=sf, textfont=sf]{subfig}
\else
\usepackage[caption=false, font=footnotesize]{subfig}
\fi

\DeclareMathOperator*{\argmax}{arg\,max}

\begin{document}

\title{Enhancing Multi-Class Classification of Random Forest using Random Vector Functional Neural Network and Oblique Decision Surfaces}

% author names and affiliations
% use a multiple column layout for up to three different
% affiliations
% \author{\IEEEauthorblockN{Rakesh Katuwal}
% \IEEEauthorblockA{School of Electrical and\\ Electronic Engineering\\
% Nanyang Technological University\\
% Singapore\\
% rakeshku001@ntu.edu.sg}
% \and
% \IEEEauthorblockN{P.N. Suganthan}
% \IEEEauthorblockA{School of Electrical and\\ Electronic Engineering\\
% Nanyang Technological University\\
% Singapore\\
% epnsugan@ntu.edu.sg}
% }

\author{\IEEEauthorblockN{Rakesh Katuwal and P.N. Suganthan}
\IEEEauthorblockA{School of Electrical and Electronic Engineering\\
Nanyang Technological University\\
Singapore\\
\{\emph{rakeshku001, epnsugan}\}@ntu.edu.sg}}

\maketitle

\begin{abstract}
Both neural networks and decision trees are popular machine learning methods and are widely used to solve problems from diverse domains. These two classifiers are commonly used base classifiers in an ensemble framework. In this paper, we first present a new variant of oblique decision tree based on a linear classifier, then construct an ensemble classifier based on the fusion of a fast neural network, random vector functional link network and oblique decision trees. Random Vector Functional Link Network has an elegant closed form solution with extremely short training time. The neural network partitions each training bag (obtained using bagging) at the root level into $C$ subsets where $C$ is the number of classes in the dataset and subsequently, $C$ oblique decision trees are trained on such partitions. The proposed method provides a rich insight into the data by grouping the confusing or hard to classify samples for each class and thus, provides an opportunity to employ fine-grained classification rule over the data. The performance of the ensemble classifier is evaluated on several multi-class datasets where it demonstrates a superior performance compared to other state-of-the-art classifiers.
\end{abstract}

\section{Introduction}

The performance of classifiers can be significantly improved by aggregating the decisions of several classifiers instead of using only a single classifier. This is generally known as ensemble of classifiers, or multiple classifier systems. The ensemble is obtained by perturbing and combining several individual classifiers \cite{breiman1996bias}. Specifically, it is obtained by perturbing the training set or injecting some randomness in each classifier and aggregating the outputs of the these classifiers in a suitable way.

Decision trees (DT) and Neural networks (NN) are generally used for ensemble generation. Both decision trees and randomized neural networks are unstable classifiers whose performance greatly vary even when there is a small perturbation in training set or some classifier parameters. Thus, they are ideal candidates for a base classifier of an ensemble framework. Random Forest (RaF) \cite{breiman2001random}, an ensemble of decision trees is an exemplar of such ensembles. It is the top ranked classifier based on the comparisons amongst 179 classifiers on 121 datasets \cite{JMLR:v15:delgado14a}. The standard random forest is however, superseded by oblique random forest, an ensemble of decision trees employing linear hyperplanes at each node to split the data instead of a single feature, in a recent exhaustive comparison among 183 classifiers \cite{zhang2017benchmarking}. An ensemble of random vector functional link (RVFL) networks, a popular single layer feed forward neural network, also ranks amongst the top-20. 

Decision trees in random forest employ recursive partitioning of the training data into smaller subsets that further aid in classification by optimizing some impurity criteria such as information gain or gini index \cite{criminisi2012decision}. Classical RaFs achieve this by using a single feature at each node to partition the training set into two partitions that generates an axis-parallel or orthogonal hyperplane at each node. Such hyperplanes may not always approximate complex decision boundaries \cite{murthy1994system, menze2011oblique, zhang2017robust}. In a variant of random forest, known as oblique random forest (obRaF) \cite{breiman1984classification}, an oblique hyperplane (or linear decision boundary) is used at each node. Such decision boundary uses a linear combination of features to split the training data. The decision trees in random forest exhaustively search for a single feature among a random subset of features at each node. However, such exhaustive search for the best oblique hyperplane is computationally expensive \cite{murthy1994system}. Thus, the search for the oblique splits are generally based on heuristic approaches and are non-optimal. To circumvent such an issue, we present an oblique random forest that searches for an optimal linear hyperplane at each node from a finite search space while optimizing the Gini-impurity criteria similar to RaF. 
  
 Random Vector Functional Link (RVFL) network on the other hand, is a randomized variant of Functional Link Neural Network (FLNN) \cite{dehuri2010comprehensive}. The weight and  bias vectors of the hidden layer in RVFL are randomly generated thus, making the learning algorithm less complicated and faster to train than conventional back-prop based SLFN \cite{pao1992neural,zhang2016comprehensive}. 
 
 It is generally cumbersome to learn from large datasets. Most of the classifiers such as random forest, support vector machine (SVM) do not scale well with datasets with large sample size, feature dimension or number of classes. Divide and conquer strategies, dimension reduction techniques are some of the common techniques employed in such cases. However, the number of classes still pose a constraint to the application of decision trees and SVM based classifiers. The approaches used to handle such scenarios trade performance with computational complexity. RVFL, on the other hand, can be effectively utilized for such large datasets. In this paper, we extend the idea proposed in \cite{katuwal2017ensemble} by fusing RVFL with our proposed oblique random forest and show that such ensembles can improve the accuracy while incurring less computational cost than random forest based ensembles. The RVFL partitions the training dataset into several subsets where confusing or difficult samples are grouped in the same subset. Such technique allows us to employ finer classification rules while focusing on confusing or difficult-to-classify samples. Through experiments on several datasets with varying sample size and number of classes, we demonstrate that our proposed oblique random forest ensemble is superior to standard random forest and its oblique variant in terms of both performance and computational requirements. We then create a hybrid ensemble of RVFL and oblique random forest to further boost the performance of the oblique random forest classifier.  

The rest of the paper is organized as follows: we present a brief review of the related works in the following section. In Section \ref{sec3}, we elucidate our approach for the hybrid ensemble. In Section \ref{sec4}, we present experimental results and comparison of our proposed hybrid ensemble with different classifiers. Finally, we present our conclusions in Section \ref{sec5}.

%-------------------------------------------------------------------------------------------
\section{Related Works}
\label{sec2}

Before proposing our hybrid ensemble classifier based on decision trees and neural networks, in this section we briefly review decision trees, random forest, random vector functional link networks and some hybridization based classification techniques.

\subsection{Decision Trees.} A decision tree consists of nodes and edges. The nodes are either internal (split) or leaf (terminal). The internal nodes are split into two child nodes until a stopping criterion is met, after which they become a leaf. Each internal node is associated with a test function defined as:

\begin{equation}
\label{eq1}
    f(\textbf{x};\Theta) = \left\{
            \begin{array}{ll}
            1 & \text{if } \textbf{x}(\Theta_{1})<\Theta_{2}\\ 
            0 & \text{otherwise}
            \end{array}
    \right.
\end{equation}

where, $\Theta_{1} \in \{1,2,\ldots,d\}$ and $\Theta_{2} \in \mathbb{R}$ is a threshold. The outcome determines the child node to which \textbf{x} is routed. For instance, 0 represents left child node while 1 represents right child node. Each node chooses the best test function $\Theta^{*}$ from a pool of potential test functions by optimizing a metric known as Gini-impurity. The objective is to make the resulting child nodes as pure as possible, i.e containing training samples of a single class only.

\begin{figure}
    \begin{center}
    \includegraphics[width=0.3\textwidth]{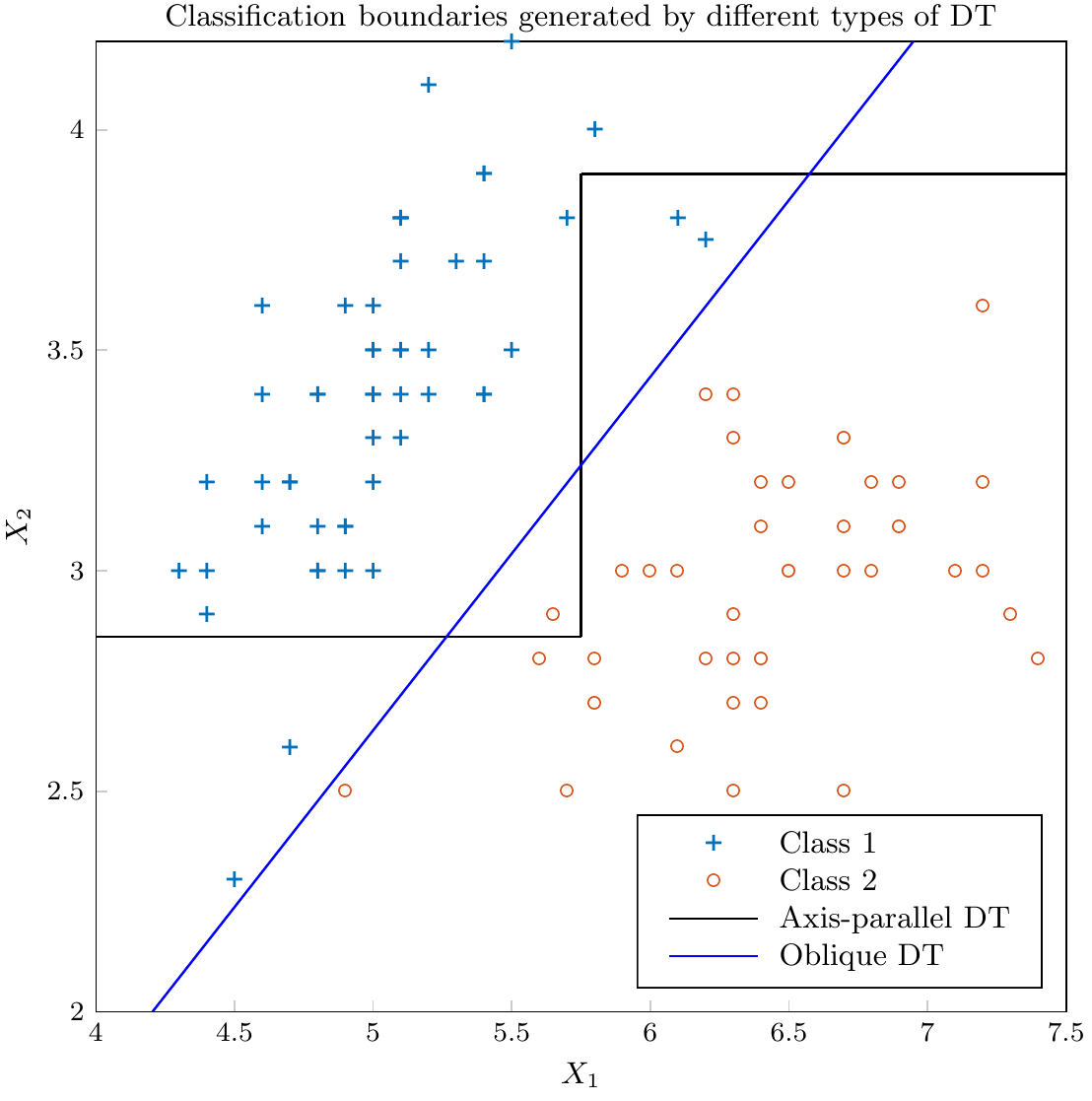}
    \caption{Classification boundaries generated by axis-parallel and oblique decision tree in a toy example for binary classification problem. Axis-parallel or staircase decision boundaries may fail to approximate the optimal decision boundary. However, oblique trees with linear hyperplanes can better perform this task.}
    \label{fig1}
    \end{center} 
\end{figure}

Based on the nature of test (split) functions, decision trees are categorized into two types: univariate (axis-parallel or orthogonal) and multivariate (oblique) \cite{breiman1984classification}. In a univariate decision tree, the parameter $\Theta$ of the test function $s(\textbf{x},\Theta)$ is based on a single feature i.e. the node selects a single feature from a random subset of features that best minimizes Gini-impurity. The final decision boundary produced by a tree is of staircase type as shown in Fig. \ref{fig1}. In multivariate or oblique decision trees, $\Theta$ depends on a linear combination of the features. Since the decision boundary can orient in any direction to the axes, the trees with such hyperplanes is also known as oblique trees \cite{breiman1984classification}. Thus, (\ref{eq1}) can be reformulated as:

\begin{equation}
    \label{eq2}
    f(\textbf{x};\Theta) = \left\{
            \begin{array}{ll}
            1 & \text{if} \sum_{i=1}^{q} w_{i}\textbf{x}_i<\Theta_{2}\\ 
            0 & \text{otherwise}
            \end{array}
    \right.
\end{equation}

where $w_i$ is the weight coefficient for each feature in $q$. 

A random forest is an ensemble of $T$ decision trees that are trained independently on several instances of the training data obtained using bagging \cite{breiman1996bagging}. In RaF, the optimal split at each node is chosen from the $n\cdot q$ possible splits, where $n$ is the unique feature values in $X_i$ and $q$ is typically set to the square root of feature dimension $d$. That means, in the worst possible scenario the exhaustive search for the `optimal' split grows linearly in the number of training samples at each node and $q$. However, there are at most $2^q \cdot {n \choose q}$ distinct oblique splits at each node and the exhaustive search of the `optimal' oblique split is NP-hard \cite{murthy1994system}. 

Because the search for optimal oblique hyperplanes is computationally expensive, many heuristic search methods such as Forest-LC \cite{breiman1984classification}, OC1 \cite{murthy1994system} have been proposed in the literature. Generally linear classifiers such as SVM, MPSVM, LDA, Logistic regression are used to generate oblique hyperplanes \cite{menze2011oblique, 6964792, zhang2007decision, lemmond2010extended, truong2009fast}. However, most of these approaches are used without optimizing impurity criteria or any search for the optimal oblique hyperplanes. A multi-class classification problem is ususally reformulated as a binary problem by defining two or more hyperclasses \cite{zhang2007decision, truong2009fast} either via clustering, some distance metric or possible combinations and a linear decision boundary is learned. These approaches either do not optimize impurity criteria or are computationally expensive. However, we can still integrate impurity optimization techniques for optimal oblique hyperplane search in ObRaF as in RaF without incuring great computational costs by implementing a simple and effective approach.

\subsection{Random Vector Functional Link Networks.}

\begin{figure}[t!]
    \begin{center}
     \includegraphics[width=\linewidth]{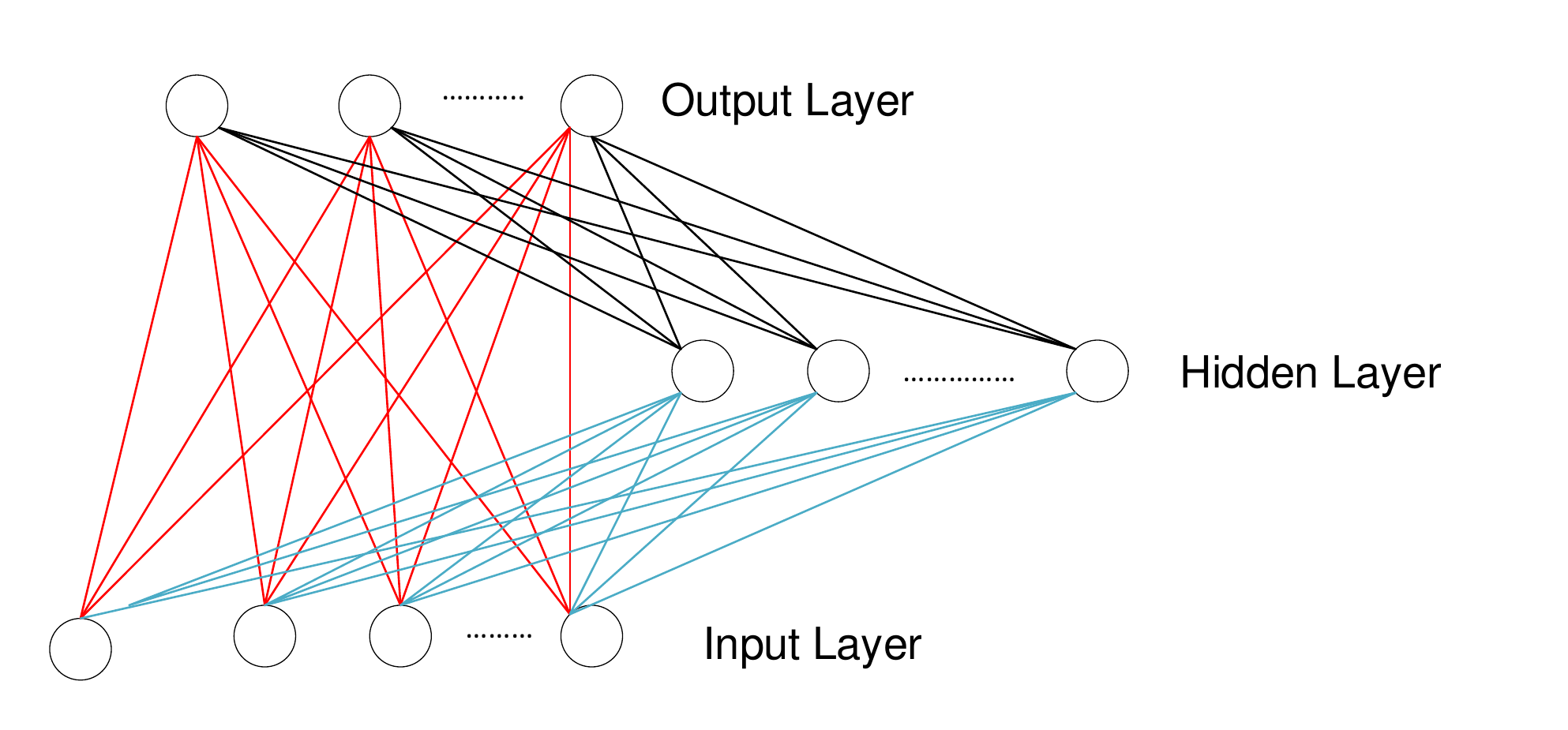}
     \caption{The structure of RVFL. The red lines are the direct links from the input to the output layer. The weights for the blue links are randomly generated from a certain range. Only the output weights needs to be computed for the red and black links. Best viewed in color.}
     \label{fig2}
    \end{center}
 \end{figure}

A RVFL is a single layer feed-forward neural network which is mainly characterized by the absence of backpropagation (BP) and the presence of direct links between the input and output nodes (see Fig. \ref{fig2}) \cite{dehuri2010comprehensive,pao1992neural}. The weights between the input and hidden neurons in RVFL are randomly generated from a suitable range. The direct links in RVFL regularize the network from the effects of randomization leading to a simpler model with a small number of hidden neurons while improving the generalization performance of the neural network \cite{zhang2016comprehensive,ren2016random}. The output layer of RVFL consists of nodes corresponding to the number of classes, with each node assigning a score for each class. The predicted class for a sample $\textbf{x}$ is the class represented by a node with $\argmax(s_i(\textbf{x}))$, $i \in \{1,\ldots,C\}$ where $s$ is the score given by each output node $i$. Since the hidden layer parameters are randomly generated and kept fixed, only the output weights need to be computed. The learning objective of RVFL is:

\begin{equation}
    \underset{w}{\textrm{min}} \phantom{i} \|Dw-Y\|^{2}+\lambda\|w\|^2
    \label{eq3}
\end{equation}
where $D = [H,X] $ is the stacked feature matrix obtained from direct links ($X$) and the hidden layer ($H$), $Y$ is the vector of class labels, and  $\lambda$ is the regularization parameter.

A closed form solution of (\ref{eq3}) can be obtained by using either least squares or Moore-Penrose pseudoinverse. Using Moore-Penrose pseudoinverse, the solution is given by: $w = D^{+}Y$ while using least squares (ridge regression), the closed form solution is given by:  
\begin{equation}
    \label{eq4}
\textrm{Primal Space:} \phantom{W} w = (D^{T}D+\lambda I)^{-1}D^{T}Y \end{equation}
\begin{equation}
    \label{eq5}
\textrm{Dual Space:} \phantom{W} w = D^{T}(DD^{T}+\lambda I)^{-1}Y \\
\end{equation}

Here, $I$ is the identity matrix. 

\subsection{Hybridization based classification techniques}
Many classifiers (heterogeneous or homogeneous) can be used as base classifiers in an ensemble framework. Here, we refer to some classification techniques employing both neural network and decision trees. In \cite{sethi1990entropy}, the authors use decision trees to empirically determine the number of neurons needed in three layers of neural network. Richmond et al. in \cite{richmond2015relating} extends this idea by mapping stacked RF to Convolutional Neural Networks (CNN) for semantic segmentation. Similarly, in \cite{jerez2003combined}, Jerez et al. use decision trees to identify most important variables from breast cancer data set and use those variables as input for their neural network architecture. The work in \cite{kontschieder2015deep} integrates NN and DT by replacing the final Softmax layer of CNN by DT. In \cite{rota2014neural}, the authors use multi-layer perceptrons as split functions in each node of the trees. Different from these works, we present a hybrid or heterogeneous ensemble of classifiers where we
exploit the probability like outputs of neural network to quickly partition the training data for efficient multi-class classification by decision trees (or random forest).

Apart from the hybridization technique discussed above, we also review some ideas relevant to our proposed method. Generally, binary splits are popular with decision trees with very few researches on multi-way splits. Multi-way (Multi-branch) splits in decision trees have previously been studied in \cite{mansour2000boosting, sadoghi2011correlation, frank1996selecting}. In \cite{sadoghi2011correlation}, correlation is used to do find the best single feature and thresholds to split the training data into multiple branches are computed by SVM. However, such multi-way splits are cumbersome to determine and do not improve the performance of decision trees \cite{katuwal2017ensemble}. \cite{murthy2016deep} is another closely related work to ours. It uses a deep neural network to perform a hierarchical partition of the data as in decision trees while creating the clusters of confusing classes. The classes are clustered by employing spectral co-clustering algorithm over confusion matrix computed over the validation dataset. Thus, it is computationally expensive and requires large dataset. However, we employ a simple, fast neural network to partition the data without incurring large computational complexities. We extend the idea of  \cite{katuwal2017ensemble} by proposing an improved oblique random forest. Our method is based on an ensemble framework that boosts the multi-class classification handling capacity of random forest. 

%-------------------------------------------------------------------------------------------
\section{RVFL and oblique random forest for many class problems} 
\label{sec3}

A RVFL followed by $C$ oblique decision trees is a base classifier in our ensemble framework. Each bag of the original training data obtained using bagging is carefully partitioned by RVFL into several subsets such that the decision trees employed afterwards can improve the classification performance by learning to separate the confusing training samples. The decision trees or more specifically ensembles of decision trees is one of the best classification algorithm in terms of generalization ability and robustness. The partitions obtained using RVFL enables to employ a more fine-grained classification rule via decision trees as the classification algorithm focuses on difficult to classify samples. In this section, we first describe the data partitioning step by RVFL and then present our oblique random forest.

\subsection{Data Partition by RVFL}

We employ RVFL at the top node to divide the data into $C$ partitions as in \cite{katuwal2017ensemble} where $C$ is the number of classes in a dataset. Our proposed oblique decision tree is trained thereafter on each partition separately to improve the accuracy. In each partition, the class distribution of samples is unique i.e. majority of the samples are from one class and the rest from other classes. The samples from the other classes are those that are ``hard'' to classify by RVFL. Such partitioning is possible by utilizing the output scores given by RVFL. 

In the training phase, each training sample $x_i$ is passed to RVFL. The output of RVFL is a probability like score for each class in that particular data set. Generally, the class with the highest score is the predicted class by RVFL. However, in our case two classes with the highest and second-highest scores are selected as the potential classes which indicate the most confusing classes for that particular training sample. Each partition by RVFL corresponds to a class. Thus, $x_i$ is used as a training data for the oblique decision trees associated with those two classes/subsets. This procedure is repeated for all training samples, creating a training set for each decision tree. The final model is an ensemble of such base classifiers. In cases, where the true class of $x_i$ is neither the highest nor the second-highest class, the training sample is still placed in its true class. For further details, readers can refer to \cite{katuwal2017ensemble}.

\subsection{Improved Oblique Random Forest}

Decision trees employ recursive partitioning so that the child nodes are purer than the parent node. The objective is to separate the training samples into different partitions such that these partitions contain samples of one class only. In RaF, such partitions are obtained by an exhaustive search for the best orthogonal hyperplane. This problem, however, can be reformulated using the information of the class labels of the training samples. 

Many popular binary classifiers such as Support Vector Machine (SVM) use ``one-vs-all" approach to breakdown a multi-class classification problem into several binary classification problems. Specifically, for each class, a single SVM is trained with the samples of that class as positive samples and all other as negative samples. A caveat is that it may not always be the best method to deal with multi-class problems \cite{991427}. Such methods can however, be integrated intactly at the internal nodes of the decision trees. As stated earlier, a linear classifier at each node need not always be a perfect classifier but simply aid in further classification. Thus, the above objective of random forest can be restated to separating one class from all other classes at each node. That means, instead of performing exhaustive search, one can search for the `$K$' hyperplanes by transforming a multi-class classification problems into `$K$' binary classification problems where `$K$' is the number of classes at each node \footnote{The total number of classes in a dataset is denoted by $C$ and the number of classes at each node by $K$. $C \geq K$ where $C=K$ at the top root node.}. This restricts the hyperplane search space in $K$ ways at each node and a linear classifier can be selected that best optimizes the impurity criteria. In the best scenario, a linear classifier or decision boundary may result in all samples of a class in one child node while the rest of the training samples on another child node which is exactly the objective of decision trees i.e. to make child nodes purer than the parent nodes. Thus, employing linear hyperplanes with impurity optimization technique may help to better capture the geometric structure of the data than axis-parallel hyperplanes.

For our oblique decision trees, we employ an MPSVM based linear classifier. MPSVM generates two non-parallel planes based on the proximity to each class and the final decision boundary employed at each node is based on the angle bisector of these two planes \cite{6964792}. The two proximal planes are found from the eigenvectors corresponding to the smallest eigenvalues of the following two generalized eigenvalue problems:

\begin{equation}
    \label{eq6}
    \begin{split}
        Gz &= \lambda Hz, z\neq0 \\
        Lz &= \lambda Mz, z\neq0 
    \end{split}
\end{equation}
where $z = [w \quad b]^{T}$, $G = [A \quad -e]^T [A \quad -e]$, 
$H = [B \quad -e]^T [B \quad -e]$ and $A$, $B$ are the matrices of class 1 and 2 respectively and $e$ is the vector of ones. The linear hyperplane at each node is the one that passes in between them.

At each node of the tree, $K$ oblique hyperplanes based on MPSVM are obtained and the one that best maximizes (\ref{eq7})  is selected as the node splitting hyperplane.

\begin{equation}
    \label{eq7}
    \textrm{max} (Gini_{p} - Gini_{c})
\end{equation}
where
\begin{equation}
    \label{eq8}
    Gini_{p} = 1-\sum_{i=1}^{K} \left(\frac{n_{w_i}}{n_{t}}\right)^2
\end{equation}

\begin{equation}
     \label{eq9}
    Gini_{c} = \frac{n_{t}^{l}}{n_{t}}\left[1-\sum_{i=1}^{K}\left (\frac{n_{w_{i}}^l}{n_{t}^{l}}\right)^2 \right] +\frac{n_{t}^{r}}{n_{t}}\left[1-\sum_{i=1}^{K} \left (\frac{n_{w_{i}}^r}{n_{t}^{r}}\right)^2 \right]
\end{equation}

where $Gini_p$ and $Gini_c$ are the values of Gini impurity at the parent node and child nodes respectively , $n_{t}$ is the number of data samples in the parent node, $n_{t}^{l}, n_{t}^{r}$ is the number of data samples that reach the left and right child nodes of the current parent node, and ${n_{w_{i}}^l}, {n_{w_{i}}^r}$ are the number of samples of class $w_{i}$ in the left and right child nodes respectively. 

The construction algorithm of the proposed oblique random forest is presented in \ref{algo1}.

\begin{algorithm}[t!]
  \caption{Proposed obRaF Training}\label{algo1}
    \begin{algorithmic}[1]
    \State \textbf{Require:} Labeled training data ($X,Y$)
    \State \textbf{Require:} Maximum tree depth $D_{max}$
    \State \textbf{Output:} A linear decision boundary ($w$ and $b$) 
       \For{\texttt{$d$ from 1 to $D_{max}$}}
         \State Check stopping criteria at each node in depth $d$.
         \State Create $K$ partitions using one-vs-all approach.
         \State Train a linear classifier for each partition.%; Eq. (\ref{eq4}) or Eq. (\ref{eq5}) or Eq. (\ref{eq6})
         %\State %Based on the training samples that fall on either side of the hyperplane 
         \State Compute Eqs. (\ref{eq8}) and (\ref{eq9}).
         \State \textbf{return} $w$ and $b$ of the hyperplane that maximizes \ref{eq7}
        \EndFor
      
    \end{algorithmic}
\end{algorithm}

One of the issues with ``one-vs-all" technique is that it is computationally voracious when the number of classes is very large. Thus, generating $K$ hyperplanes at each node can be computationally expensive in such cases. However, it also offers an advantage. Since the linear hyperplane at each node is selected from a pool of hyperplanes based on the impurity criteria (more specifically (\ref{eq7})), it results in pure child nodes faster than previous exhaustive approaches (in case of RaF) and non-impurity optimization approaches (in case of obRaF) \footnote{When we say obRaF, we are referring to the older variants of oblique random forest that do not employ impurity optimization techniques.}. Thus, the trees in our proposed oblique random forest variant are generally shallow compared to the standard trees. This may negate the complexity associated with generating many hyperplanes at each node. In Section \ref{sec4}, we validate this through experiments on several datasets. 

When employing RVFL at the top (root) node, each subset or partition contains majority of the samples from few classes and rest from the others. Because the linear classifiers trained with greater number of training samples are most likely to better optimize the impurity (Gini) criteria compared to the classifiers trained with very few samples, we can avoid such classes and instead choose the best hyperplane from a pool of hyperplanes obtained using classes with larger training data only. This is intuitive since the hyperplane generated using ``one-vs-all" method attempts to separate one class from the rest of the other classes and thus, it favours classes with larger training data as it better optimizes (\ref{eq7}). This observation is also based on our experiments where the best hyperplane is usually the one trained with larger number of classes. Thus, when the number of classes is very large, we employ the ``one-vs-all" approach to only the top $k\%$ classes where $k$ is a hyperparamater. Based on our experiments, we set $k=50$. This further decreases the computational complexity of the model without incuring any significant loss in the performance. 

Our method is particularly suitable for multi-core or distributed environment where after the partitioning by RVFL, each partition can be run in parallel or distributed across different cores. Similarly, the hyperplane generation operation using ``one-vs-all" can also be distributed. This method is also suitable for large datasets. However, a caveat associated with our ensemble classifier is that it can only be employed for many class classification problems. For binary classes, the partitions provided by RVFL are just duplicated versions.
%--------------------------------------------------------------------------
\section{Experiments}
\label{sec4}
In this section, we compare the performance of random forests variants and our hybrid ensemble. We compare four classifiers in 10 UCI multi-class datasets. The number of classes in these datasets vary from 7 to 100. The datasets are selected based on their size and number of classes and the performance of oblique random forest and the hybrid ensemble in \cite{katuwal2017ensemble}. 
The properties of these datasets are summarized in Table \ref{Table1}.

\begin{threeparttable}
    \caption{Overview of the UCI datasets}
    \label{Table1} 
    \begin{tabular}{l c c c c} \toprule
        Dataset & \#Patterns & \#Features & \#Classes \\ \midrule
        Chess-krvk & 28056 & 6 & 18\\
        Letter & 20000 & 16 & 26 \\
        Optical & 3823 & 62 & 10 \\
        Pendigits & 7494 & 16 & 10 \\
        Plant margin & 1600 & 64 & 100 \\
        Plant shape & 1600 & 64 & 100 \\
        Statlog-image & 2310 & 18 & 7 \\
        USPS & 9298 & 252 & 10 \\ 
        W-qua-white & 4898 & 11 & 7 \\
        Yeast & 1484 & 8 & 10 \\
        \bottomrule
    \end{tabular}
        
\end{threeparttable}

\subsection{Experimental setup}
We follow the experimental setup of \cite{JMLR:v15:delgado14a, katuwal2017ensemble}. For a fair comparison between all the ensemble methods, we use the same values for the common parameters. Thus, for each classifier, we set the ensemble size or the number of trees to 500, number of random features at each node ($q$) to $\sqrt d$, where $d$ is the feature dimension. If the feature vector has not been normalized, each feature is normalized by removing the mean and dividing by its variance. In all the ensembles, the trees are fully grown until the terminating criteria is met (no longer possible to optimize \ref{eq7}). 
\singlespacing
\textbf{RVFL Configuration.}
The objective of our proposed method is to obtain diverse RVFL models in each base classifier which in turn results in diverse data partitions hence, diverse decision trees. We use the same parameter setting as in \cite{katuwal2017ensemble} where each RVFL randomly picks the activation function and network parameters from the parameter settings listed below:

\begin{enumerate}
\item Number of hidden neurons, $\mathcal{N}$ = 3:203
\item $\lambda$ (=$(1/2)^C$) in ridge regression, $C$ = -5:14
\item Activation Functions: \textit{radbas}, \textit{sine} and \textit{tribas}
\item Range of the randomization for weights [-$S$,+$S$] and bias [0,$S$], where $S = 2^t$ with t = -1.5:0.5:1.5
\end{enumerate}

The RVFL has direct links from input layer to output layer with bias term in the output neuron.

\subsection{Comparison between random forest variants}
In Table \ref{Table4}, we present the classification accuracies of each classifier in each dataset. First we compare the performance of random forest variants: RaF, obRaF and obRaF(M). In almost all the datasets except Plant margin, our proposed oblique random forest (obRaF(M)) outperforms both standard random forest and oblique random forest. Although both obRaF and obRaF(M) employ linear decision boundary at each node using MPSVM, only obRaF(M) performs a search for the optimal linear boundary. This suggests that oblique random forests that employ linear decision boundaries with impurity optimization generalize better than other random forest variants.

In Tables \ref{Table2} and \ref{Table3}, we show the training time and the average number of nodes comparison of random forest classifiers. For the comparison, we select two datasets: Plant shape with 100 classes and Pendigits, a medium size dataset. Even though our proposed oblique random forest employs ``one-vs-all" approach, it still offers computational advantages over RaF and obRaF which is evident by shorter training time and less number of nodes.

\begin{table*}[t!] 
    \begin{center}
    \begin{threeparttable}
    \caption{Training time comparison (in seconds) between random forest classifiers}
    \label{Table2} 
        \begin{tabular}{l c c c}\toprule
            Dataset & RaF & obRaF & obRaF(M) \\ \midrule
             Plant shape & 656.64 & 678.99 & 622.21 \\
             Pendigits  & 422.24 & 357.51 & 352.17 \\ \bottomrule 
        \end{tabular}
    \end{threeparttable}
    \end{center}
    
\end{table*}

\begin{table*}[t!] 
    \begin{center}
    \begin{threeparttable}
    \caption{Average number of nodes in the random forest classifiers}
    \label{Table3} 
        \begin{tabular}{l c c c}\toprule
            Dataset & RaF & obRaF & obRaF(M) \\ \midrule
             Plant shape & 291.74 & 331.78 & 264.01 \\
             Pendigits  & 307.78 & 301.36 & 290.92 \\ \bottomrule 
        \end{tabular}
    \end{threeparttable}
    \end{center}
    
\end{table*}

\subsection{Performance comparison between all classifiers}
The average classification accuracies (in \%) of RaF, obRaF, obRaF(M) and obRaFL are 83.07, 83.68, 84.36 and 84.6 respectively. The hybrid ensemble (obRaFL) has the highest accuracy followed by our proposed oblique random forest, obRaF(M). However, comparing the classifiers using average accuracy is susceptible to outliers and may atone for a classifier's poor performance in one dataset with an excellent performance on the other. Thus, we follow the procedure of \cite{JMLR:v15:delgado14a}, and use the rank of each classifier to assess its performance. In this approach, each classifier is ranked based on its performance, that means, the highest performing classifier is ranked 1, the second highest rank 2, and so on. The mean ranks of each classifier over all the datasets is presented in Table \ref{Table3}. Our hybrid ensemble, obRaFL, is the top ranked classifier followed by our proposed oblique random forest, obRaF(M). 

Thus, from the experimental results, we can infer that employing obRaF(M) on the partitions provided by RVFL improves the performance. The RVFL provides partitions with confusing samples so classification rules that focus on such difficult-to-classify samples can be employed. Such rules can be easily implemented by decision trees or random forest owing to their superior and robust performance. Furthermore, our proposed oblique random forest improves the random forest and by employing RVFL at the top node, we can obtain more robust and superior performance.

\begin{table*}[ht!]
\begin{center}
    \begin{threeparttable}
        \caption{Accuracies (\%) of different Random Forest based methods} 
        \label{Table4} 
            \begin{tabular}{l c c c c} \toprule
                Dataset & RaF & obRaF & obRaF(M) & obRaFL \\ \midrule
                Chess-krvk & 70.48 & 68.19  & 74.35 & \textbf{75.06} \\
                Letter & 96 & 96.53 & 97.02 & \textbf{97.58} \\
                Optical & 97.16 & 96.72 & 97.08 & \textbf{97.27} \\
                Pendigits & 95.31 & 96.94  & 97.13 & \textbf{97.15} \\
                Plant margin & \textbf{85.5} & 82.82 & 82.92 & 82.98\\
                Plant shape & 64.06 & 70.25 & 70.56 & \textbf{70.87} \\
                Statlog-image & 97.66 & 97.53 & 98.05 & \textbf{98.27} \\
                USPS & 93.54 & 93.55 & 93.87 & \textbf{93.94} \\ 
                W-qua-white & 68.38 &  69.24 & 69.26 & \textbf{69.49} \\
                Yeast & 62.67 & 63.14 & 63.41 & \textbf{63.48} \\
                \hline
                \textbf{Mean Acc.} & 83.07 & 83.68 & 84.36 & \textbf{84.6} \\
                 \bottomrule
            \end{tabular}
                \begin{tablenotes}
                    \small
                    \item obRaF is the oblique random forest of \cite{6964792}. obRaF(M) is our proposed oblique random forest while obRaFL is the hybrid ensemble of RVFL and obRaF(M). Bold values indicate the best performance.
                \end{tablenotes}
    \end{threeparttable}
\end{center}
\end{table*}

\begin{table*}
    \begin{center}
    \begin{threeparttable}
    \caption{Average Friedman rank based on classification accuracies of each method} 
    \label{Table5}
        \begin{tabular}{l c c c c} \toprule
            & RaF & obRaF & obRaF(M) & obRaFL\\ \midrule
            Rank &  3.3 & 3.4 & 2.2 & \textbf{1.1}\\ \bottomrule
        \end{tabular}
        \begin{tablenotes}
            \small
                \item Lower rank reflects better performance.
        \end{tablenotes}
    \end{threeparttable}
    \end{center}
\end{table*}

\subsection{Analysis of Common Parameters}
\begin{figure} 
    \centering
  \subfloat[]{%
       \includegraphics[width=0.5\linewidth]{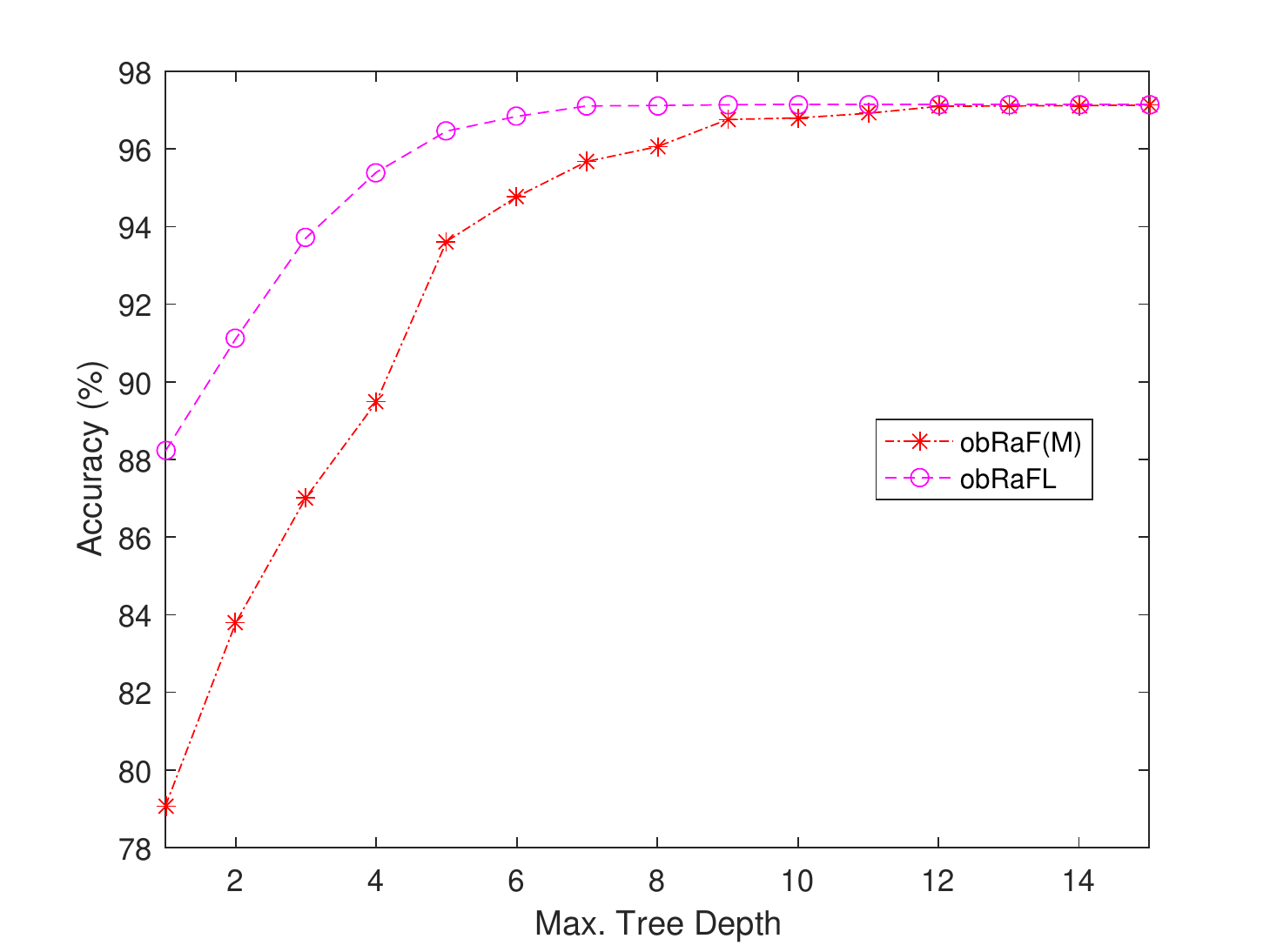}}
    \hfill
  \subfloat[]{%
        \includegraphics[width=0.5\linewidth]{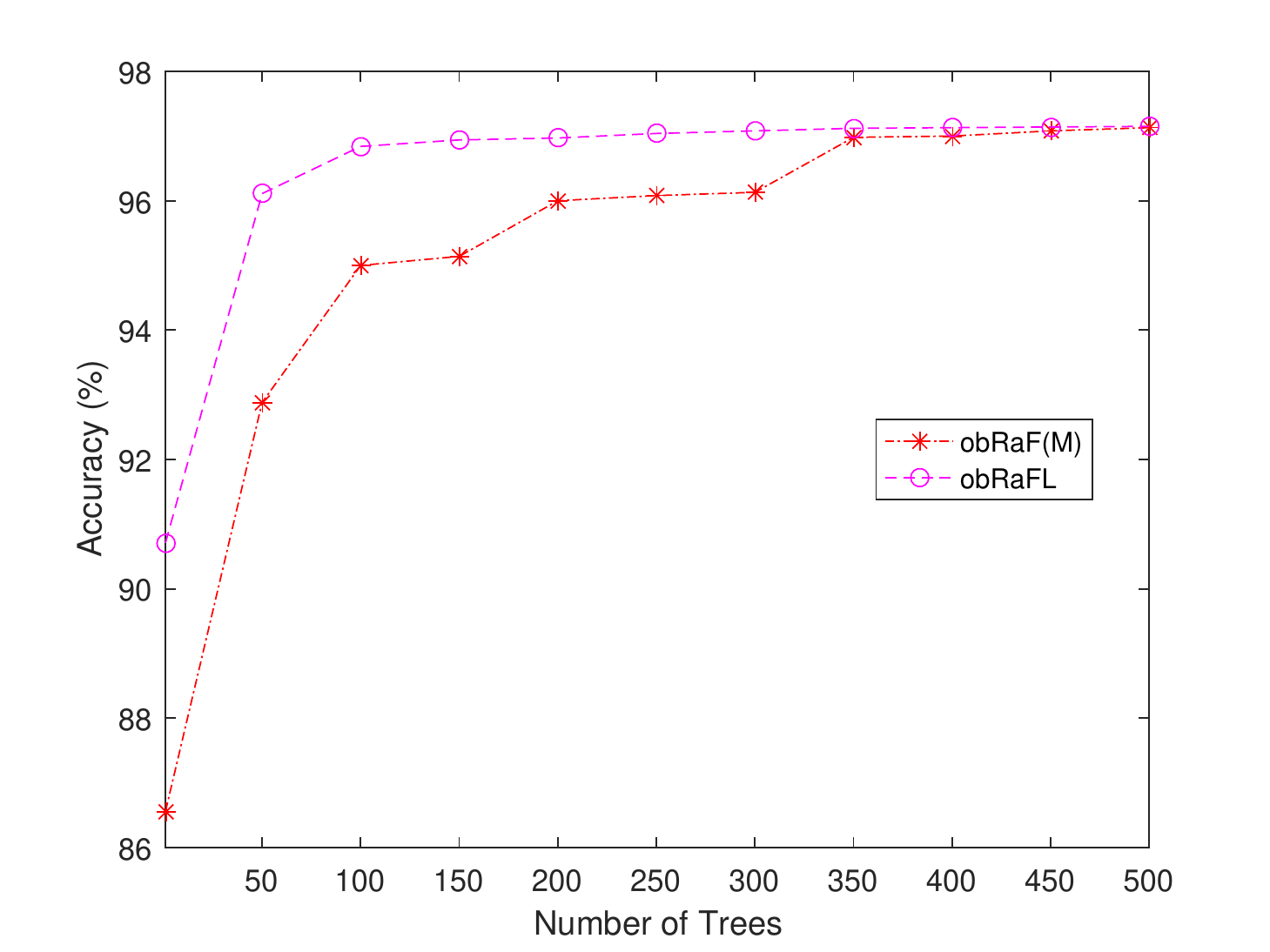}}
     \\
    \subfloat[]{%
          \includegraphics[width=0.52\linewidth]{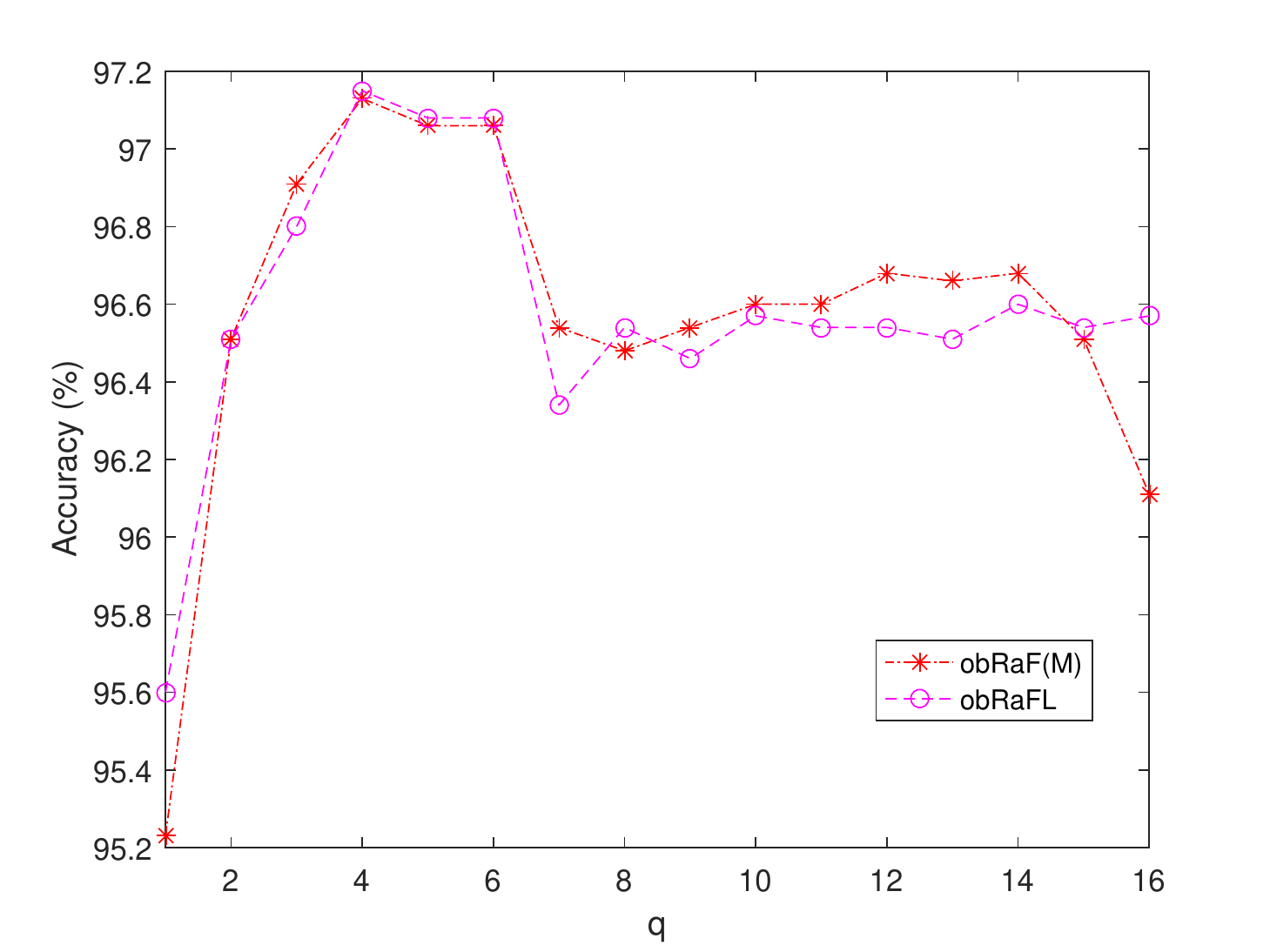}}
          
  \caption{Influence of (a) the maximum depth (b) the number of trees and (c) q on our proposed oblique random forest and hybrid classifiers.}
  \label{fig3} 
\end{figure}

Tree depth, the number of features randomly selected at each node and the number of trees are the common parameters of random forest based methods. We evaluate the influence of each parameter in our proposed oblique random forest and the hybrid classifier in the Pendigits dataset. Similar conclusions pertain to other datasets as well. For the analysis, the maximum depth, the number of trees and the number of random features are varied in the ranges [1,15], [1,500] and [1,16] respectively.
\singlespacing
\textbf{Tree depth.} Generally, the trees in random forest are fully grown. From Fig. \ref{fig3}, we can see that there is a sharp increase in the accuracy when the tree depth increases from 1 to 6. This implies that to obtain a good performance, the trees should be deeper. As the trees in the hybrid classifier are grown with reduced data set, the trees are usually shallow. However, it gives good performance than oblique random forest even when $D_{max}$ is set to a small value because of the nature of the data partitions (small size and fine-grained classification rules).

\singlespacing
\textbf{Number of features.} The number of features randomly selected at each node, $q$, controls the diversity of the trees in forest. A small value of $q$ results in uncorrelated trees whereas large values of $q$ may result in correlated trees. Generally, $q = \sqrt d$ gives good performance.
\singlespacing
\textbf{Number of trees.} As the number of trees (base classifiers) increases, the generalization ability of random forest based methods also increases. However, large number of trees or ensemble size also increases the computational cost. There is only a slight improvement on the performance beyond the ensemble size of 300.
\singlespacing
In Fig. \ref{fig3}, we can observe that even for small tree depth and ensemble size, our hybrid ensemble classifier provides the near maximal performance. For large datasets, random forest require a large number of very deep trees to provide an acceptable performance. This may be computationally intractable. However, by employing our hybrid ensemble with shallow trees and small ensemble size, we can eschew expensive computational requirements and still obtain good performance.

%-----------------------------------------------------------------------------------------
\section{Conclusion}
\label{sec5}
In this paper, we first proposed an oblique decision tree that uses impurity optimization techniques similar to the trees in random forests. We employed the oblique decision trees with a fast RVFL network to create a hybrid ensemble. In each base classifier, the decision trees were trained on the samples partitioned by RVFL. Such a marriage of decision trees and fast neural network further enhances the capability of decision trees to handle multi-class classification problem which is evident by the performance of the hybrid ensemble in several machine learning datasets. Even with small tree depth and ensemble size, our hybrid ensemble can achieve superior performance compared to standard random forest classifiers. This can significantly preserve the computational resources and time when dealing with large datasets. One of the interesting traits of our hybrid ensemble is parallelization or distributed computation as several jobs can be effectively distributed over several cores or machines. However, the gain can only be seized if we  reduce the communication overhead. Even though the proposed oblique random forest has faster training time, the partitioning and the application of RVFL adds complexity to the ensemble. Thus, our future work is to improve the training time of our hybrid ensemble with efficient use of distributed computing.

\bibliographystyle{IEEEtran}
\bibliography{ref2}

% Generated by IEEEtran.bst, version: 1.14 (2015/08/26)
\begin{thebibliography}{10}
\providecommand{\url}[1]{#1}
\csname url@samestyle\endcsname
\providecommand{\newblock}{\relax}
\providecommand{\bibinfo}[2]{#2}
\providecommand{\BIBentrySTDinterwordspacing}{\spaceskip=0pt\relax}
\providecommand{\BIBentryALTinterwordstretchfactor}{4}
\providecommand{\BIBentryALTinterwordspacing}{\spaceskip=\fontdimen2\font plus
\BIBentryALTinterwordstretchfactor\fontdimen3\font minus
  \fontdimen4\font\relax}
\providecommand{\BIBforeignlanguage}[2]{{%
\expandafter\ifx\csname l@#1\endcsname\relax
\typeout{** WARNING: IEEEtran.bst: No hyphenation pattern has been}%
\typeout{** loaded for the language `#1'. Using the pattern for}%
\typeout{** the default language instead.}%
\else
\language=\csname l@#1\endcsname
\fi
#2}}
\providecommand{\BIBdecl}{\relax}
\BIBdecl

\bibitem{breiman1996bias}
L.~Breiman, ``Bias, variance, and arcing classifiers,'' \emph{Tech. Rep. 460,
  Statistics Department, University of California, Berkeley, CA, USA}, 1996.

\bibitem{breiman2001random}
------, ``Random forests,'' \emph{Machine learning}, vol.~45, no.~1, pp. 5--32,
  2001.

\bibitem{JMLR:v15:delgado14a}
\BIBentryALTinterwordspacing
M.~Fern\'{a}ndez-Delgado, E.~Cernadas, S.~Barro, and D.~Amorim, ``Do we need
  hundreds of classifiers to solve real world classification problems?''
  \emph{Journal of Machine Learning Research}, vol.~15, pp. 3133--3181, 2014.
  [Online]. Available: \url{http://jmlr.org/papers/v15/delgado14a.html}
\BIBentrySTDinterwordspacing

\bibitem{zhang2017benchmarking}
L.~Zhang and P.~N. Suganthan, ``Benchmarking ensemble classifiers with novel
  co-trained kernal ridge regression and random vector functional link
  ensembles [research frontier],'' \emph{IEEE Computational Intelligence
  Magazine}, vol.~12, no.~4, pp. 61--72, 2017.

\bibitem{criminisi2012decision}
A.~Criminisi, J.~Shotton, E.~Konukoglu \emph{et~al.}, ``Decision forests: A
  unified framework for classification, regression, density estimation,
  manifold learning and semi-supervised learning,'' \emph{Foundations and
  Trends{\textregistered} in Computer Graphics and Vision}, vol.~7, no. 2--3,
  pp. 81--227, 2012.

\bibitem{murthy1994system}
S.~K. Murthy, S.~Kasif, and S.~Salzberg, ``A system for induction of oblique
  decision trees,'' \emph{Journal of artificial intelligence research}, vol.~2,
  pp. 1--32, 1994.

\bibitem{menze2011oblique}
B.~H. Menze, B.~M. Kelm, D.~N. Splitthoff, U.~Koethe, and F.~A. Hamprecht, ``On
  oblique random forests,'' in \emph{Joint European Conference on Machine
  Learning and Knowledge Discovery in Databases}.\hskip 1em plus 0.5em minus
  0.4em\relax Springer, 2011, pp. 453--469.

\bibitem{zhang2017robust}
L.~Zhang, J.~Varadarajan, P.~N. Suganthan, N.~Ahuja, and P.~Moulin, ``Robust
  visual tracking using oblique random forests,'' in \emph{IEEE International
  Conference on Computer Vision and Pattern Recognition. IEEE}, 2017.

\bibitem{breiman1984classification}
L.~Breiman, J.~Friedman, C.~J. Stone, and R.~A. Olshen, \emph{Classification
  and regression trees}.\hskip 1em plus 0.5em minus 0.4em\relax CRC press,
  1984.

\bibitem{dehuri2010comprehensive}
S.~Dehuri and S.-B. Cho, ``A comprehensive survey on functional link neural
  networks and an adaptive pso--bp learning for cflnn,'' \emph{Neural Computing
  and Applications}, vol.~19, no.~2, pp. 187--205, 2010.

\bibitem{pao1992neural}
Y.-H. Pao, S.~M. Phillips, and D.~J. Sobajic, ``Neural-net computing and the
  intelligent control of systems,'' \emph{International Journal of Control},
  vol.~56, no.~2, pp. 263--289, 1992.

\bibitem{zhang2016comprehensive}
L.~Zhang and P.~N. Suganthan, ``A comprehensive evaluation of random vector
  functional link networks,'' \emph{Information sciences}, vol. 367, pp.
  1094--1105, 2016.

\bibitem{katuwal2017ensemble}
R.~Katuwal, P.~Suganthan, and L.~Zhang, ``An ensemble of decision trees with
  random vector functional link networks for multi-class classification,''
  \emph{Applied Soft Computing}, 2017.

\bibitem{breiman1996bagging}
L.~Breiman, ``Bagging predictors,'' \emph{Machine learning}, vol.~24, no.~2,
  pp. 123--140, 1996.

\bibitem{6964792}
L.~Zhang and P.~N. Suganthan, ``Oblique decision tree ensemble via multisurface
  proximal support vector machine,'' \emph{IEEE Transactions on Cybernetics},
  vol.~45, no.~10, pp. 2165--2176, Oct 2015.

\bibitem{zhang2007decision}
L.~Zhang, W.-D. Zhou, T.-T. Su, and L.-C. Jiao, ``Decision tree support vector
  machine,'' \emph{International Journal on Artificial Intelligence Tools},
  vol.~16, no.~01, pp. 1--15, 2007.

\bibitem{lemmond2010extended}
T.~D. Lemmond, B.~Y. Chen, A.~O. Hatch, and W.~G. Hanley, ``An extended study
  of the discriminant random forest,'' in \emph{Data Mining}.\hskip 1em plus
  0.5em minus 0.4em\relax Springer, 2010, pp. 123--146.

\bibitem{truong2009fast}
A.~K.~Y. Truong, ``Fast growing and interpretable oblique trees via logistic
  regression models,'' Ph.D. dissertation, University of Oxford, 2009.

\bibitem{ren2016random}
Y.~Ren, P.~N. Suganthan, N.~Srikanth, and G.~Amaratunga, ``Random vector
  functional link network for short-term electricity load demand forecasting,''
  \emph{Information Sciences}, vol. 367, pp. 1078--1093, 2016.

\bibitem{sethi1990entropy}
I.~K. Sethi, ``Entropy nets: from decision trees to neural networks,''
  \emph{Proceedings of the IEEE}, vol.~78, no.~10, pp. 1605--1613, 1990.

\bibitem{richmond2015relating}
D.~L. Richmond, D.~Kainmueller, M.~Y. Yang, E.~W. Myers, and C.~Rother,
  ``Relating cascaded random forests to deep convolutional neural networks for
  semantic segmentation,'' \emph{arXiv preprint arXiv:1507.07583}, 2015.

\bibitem{jerez2003combined}
J.~M. Jerez-Aragon{\'e}s, J.~A. G{\'o}mez-Ruiz, G.~Ramos-Jim{\'e}nez,
  J.~Mu{\~n}oz-P{\'e}rez, and E.~Alba-Conejo, ``A combined neural network and
  decision trees model for prognosis of breast cancer relapse,''
  \emph{Artificial intelligence in medicine}, vol.~27, no.~1, pp. 45--63, 2003.

\bibitem{kontschieder2015deep}
P.~Kontschieder, M.~Fiterau, A.~Criminisi, and S.~Rota~Bulo, ``Deep neural
  decision forests,'' in \emph{Proceedings of the IEEE International Conference
  on Computer Vision}, 2015, pp. 1467--1475.

\bibitem{rota2014neural}
S.~Rota~Bulo and P.~Kontschieder, ``Neural decision forests for semantic image
  labelling,'' in \emph{Proceedings of the IEEE Conference on Computer Vision
  and Pattern Recognition}, 2014, pp. 81--88.

\bibitem{mansour2000boosting}
Y.~Mansour and D.~A. McAllester, ``Boosting with multi-way branching in
  decision trees,'' in \emph{Advances in Neural Information Processing
  Systems}, 2000, pp. 300--306.

\bibitem{sadoghi2011correlation}
H.~Sadoghi~Yazdi, N.~Salehi~Moghaddami, and H.~Poostchi~Mohammadabadi,
  ``Correlation based splitting criterionin multi branch decision tree,''
  \emph{Central European Journal of Computer Science}, vol.~1, 2011.

\bibitem{frank1996selecting}
E.~Frank and I.~H. Witten, ``Selecting multiway splits in decision trees,''
  1996.

\bibitem{murthy2016deep}
V.~N. Murthy, V.~Singh, T.~Chen, R.~Manmatha, and D.~Comaniciu, ``Deep decision
  network for multi-class image classification,'' in \emph{Computer Vision and
  Pattern Recognition (CVPR), 2016 IEEE Conference on}.\hskip 1em plus 0.5em
  minus 0.4em\relax IEEE, 2016, pp. 2240--2248.

\bibitem{991427}
C.-W. Hsu and C.-J. Lin, ``A comparison of methods for multiclass support
  vector machines,'' \emph{IEEE Transactions on Neural Networks}, vol.~13,
  no.~2, pp. 415--425, Mar 2002.

\end{thebibliography}

% that's all folks
\end{document}